%% file: main.tex
\documentclass[11pt,a4paper]{article}
\usepackage{times,latexsym}
\usepackage{url}
\usepackage[T1]{fontenc}

\usepackage[]{gecko2020v1}

\usepackage{graphicx} 

\title{Novel Aficionados and Doppelg\"{a}ngers: a referential task for semantic representations of individual entities}

\author{Andrea Bruera \\
  Queen Mary University of London\\
  School of Electronical Engineering and\\
  Computer Science \\
  \texttt{a.bruera@qmul.ac.uk} \\\And
  Aurélie Herbelot \\
  University of Trento\\
  Center for Mind and \\
  Brain Sciences \\
  }

\aclfinalcopy

\begin{document}
	\maketitle

\input{abstract.tex}

\begin{figure*}[ht]
    \centering
    \includegraphics[width=\textwidth]{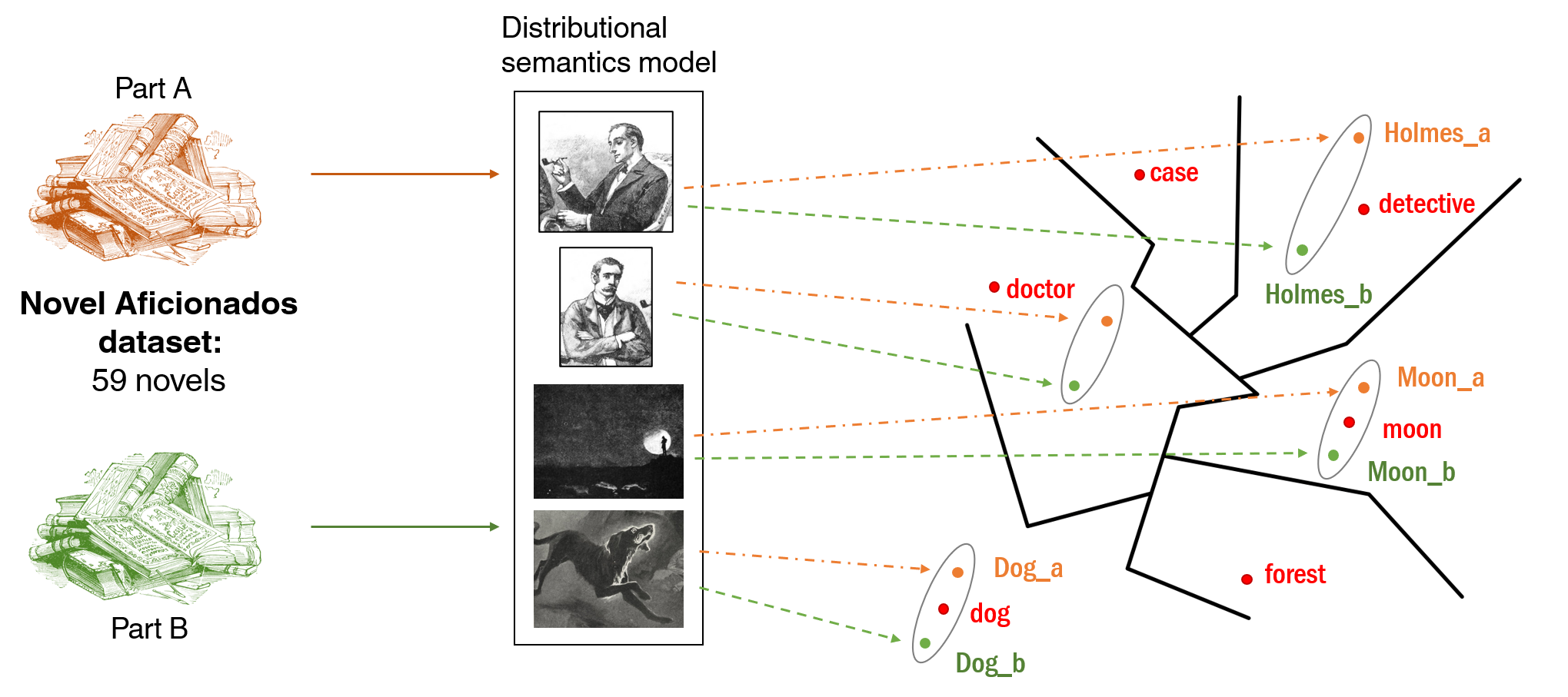}
    \caption{A visualization of the Doppelg\"{a}nger test.
    Each of the 59 novels is split into two parts (Part A and Part B), and then from each one of them, for each character and for the matched common nouns, a word vector is created by using distributional semantics models. Then, by comparing the vectors for part A and part B, we check whether we can correctly match co-referring word vectors.}
    \label{fig:dop}
\end{figure*}

\section{Introduction}
\input{introduction.tex}
\section{Related work}
\input{related_work.tex}

\begin{figure*}[ht]
    \centering
    \includegraphics[width=\textwidth]{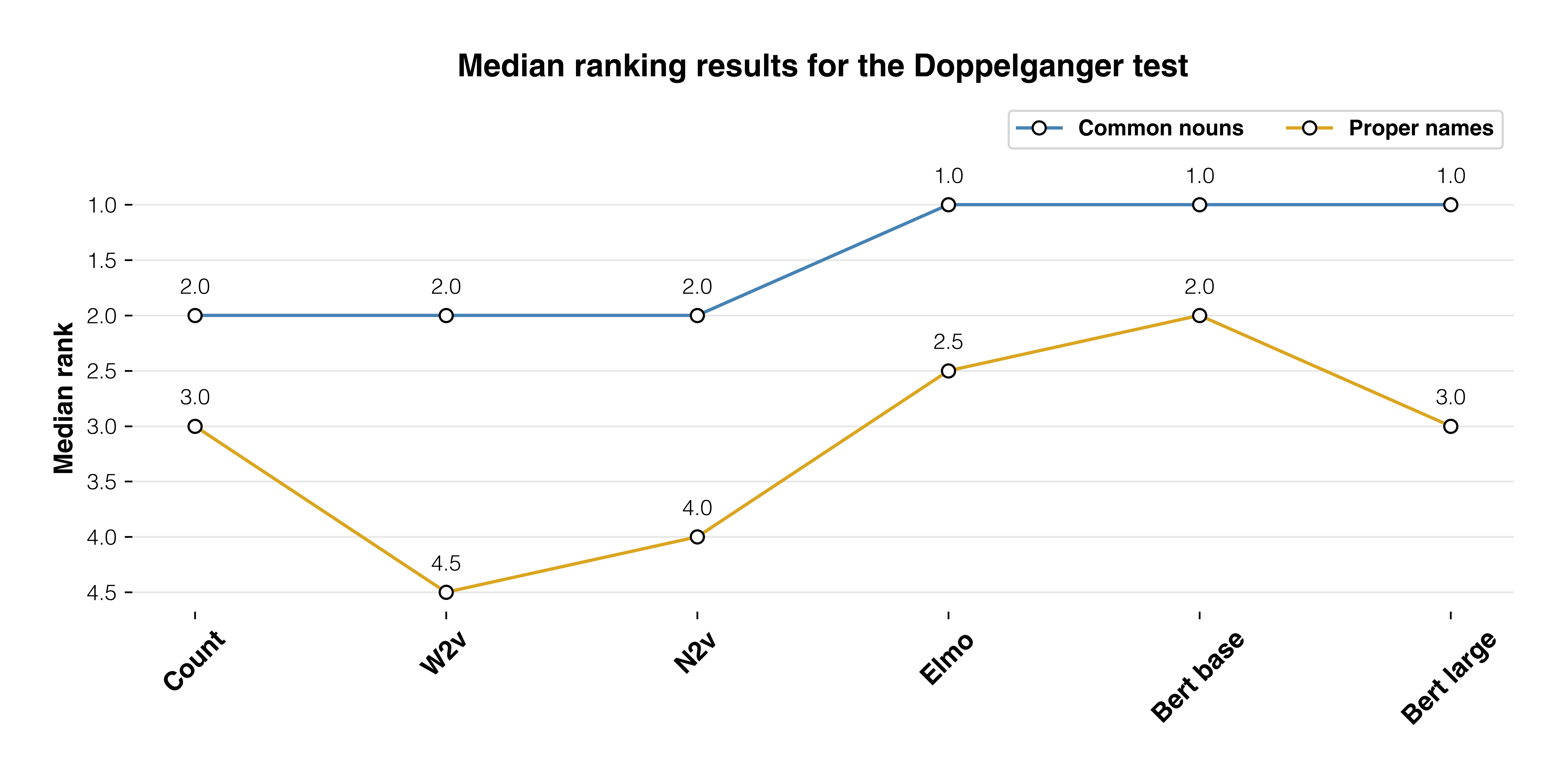}
    \caption{Results for the Doppelg\"{a}nger test.}
    \label{fig:dop_results}
\end{figure*}

\section{Data}
\input{data.tex}

\begin{figure*}[ht]
    \centering
    \includegraphics[width=\textwidth]{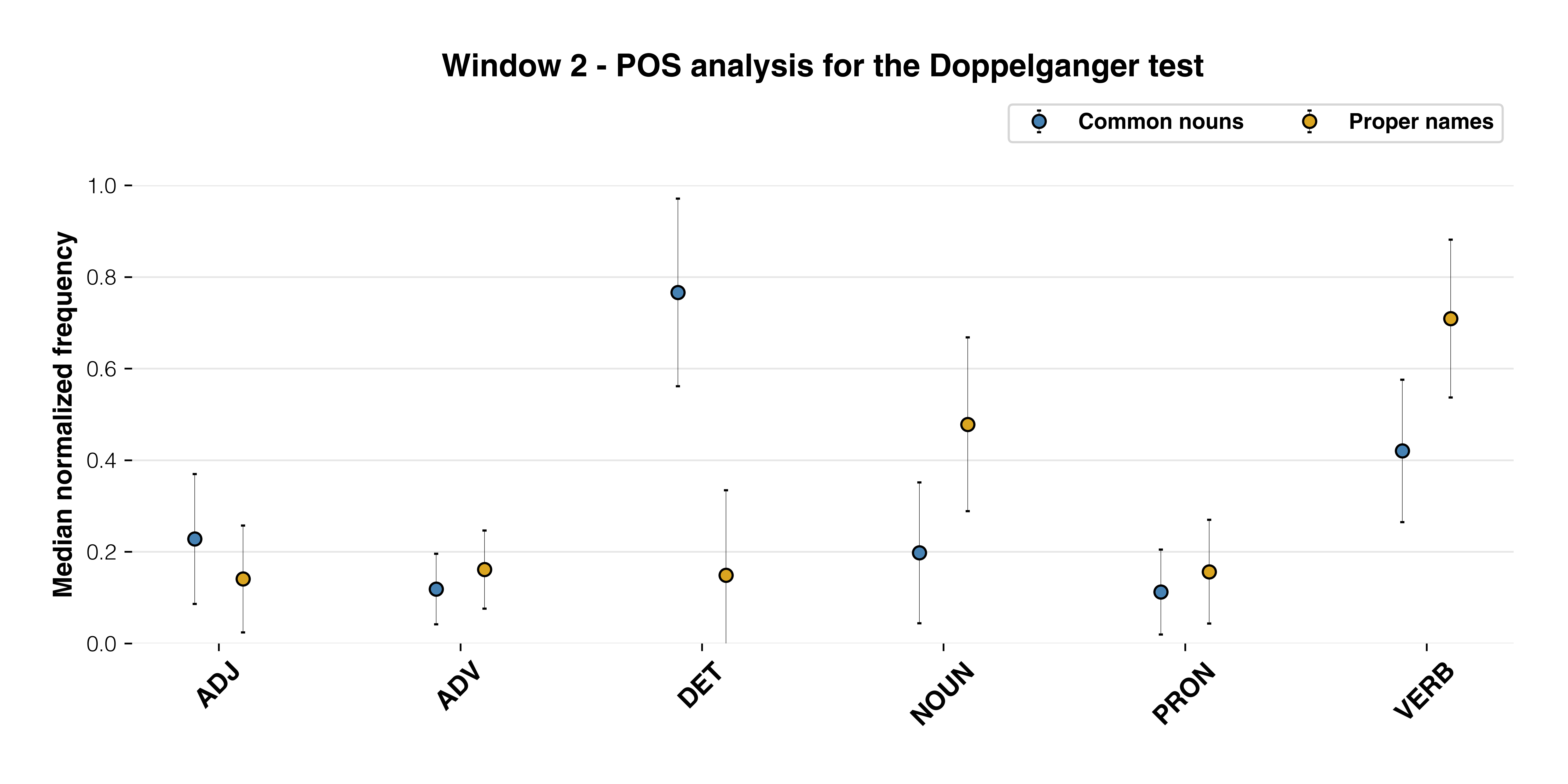}
    \caption{Results for the part-of-speech analyses, indicating differences among distributions of parts of speech around proper names and common nouns with respect to nouns and verbs (proper names > common nouns).}
    \label{fig:pos}
\end{figure*}

\section{Task}
\input{task.tex}

\begin{figure*}[ht]
    \centering
    \includegraphics[width=\textwidth]{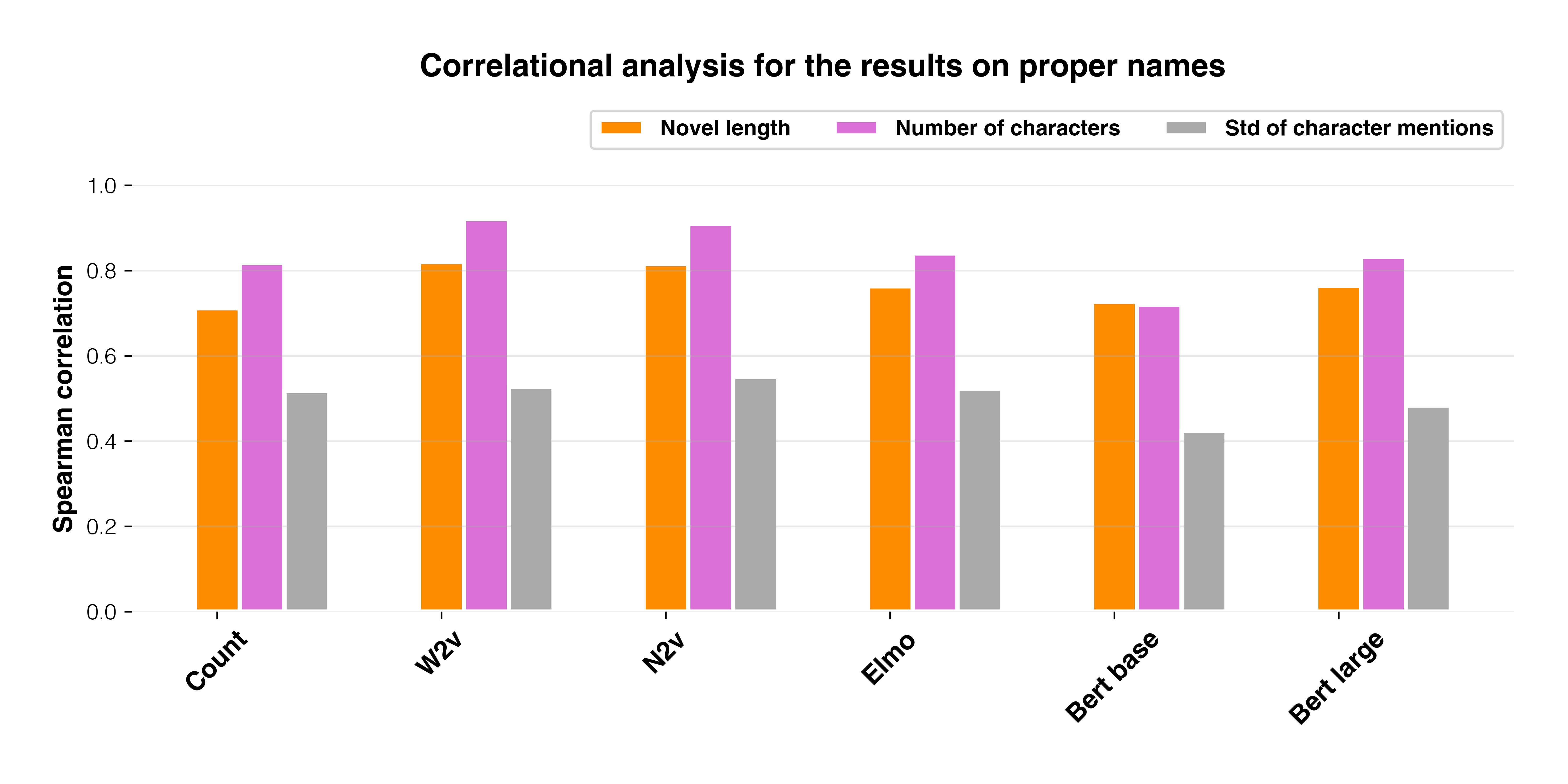}
    \caption{Results for the correlational analyses across all models.}
    \label{fig:correlations}
\end{figure*}

\section{Models}
\input{models.tex}

\begin{figure*}[ht]
    \centering
    \includegraphics[width=\textwidth]{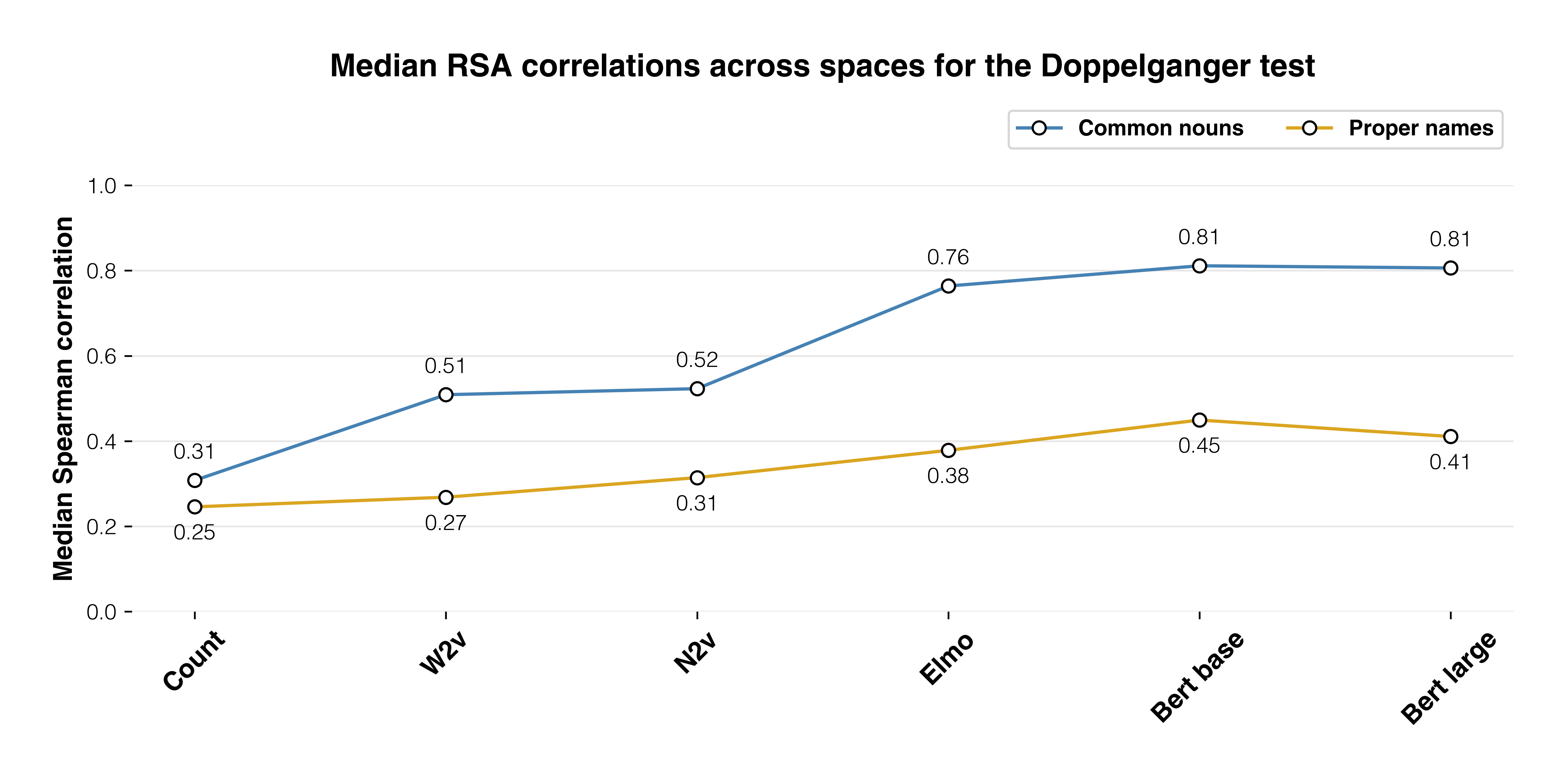}
    \caption{Results for the Representational Similarity Analysis. Vector spaces across parts of the novels are more similar in the case of common nouns than in the case of proper names - confirming that it proper names pose peculiar challenges to distributional semantic models.}
    \label{fig:rsa}
\end{figure*}

\section{Results}

\input{results.tex}

\begin{figure*}[ht]
    \centering
    \includegraphics[width=\textwidth]{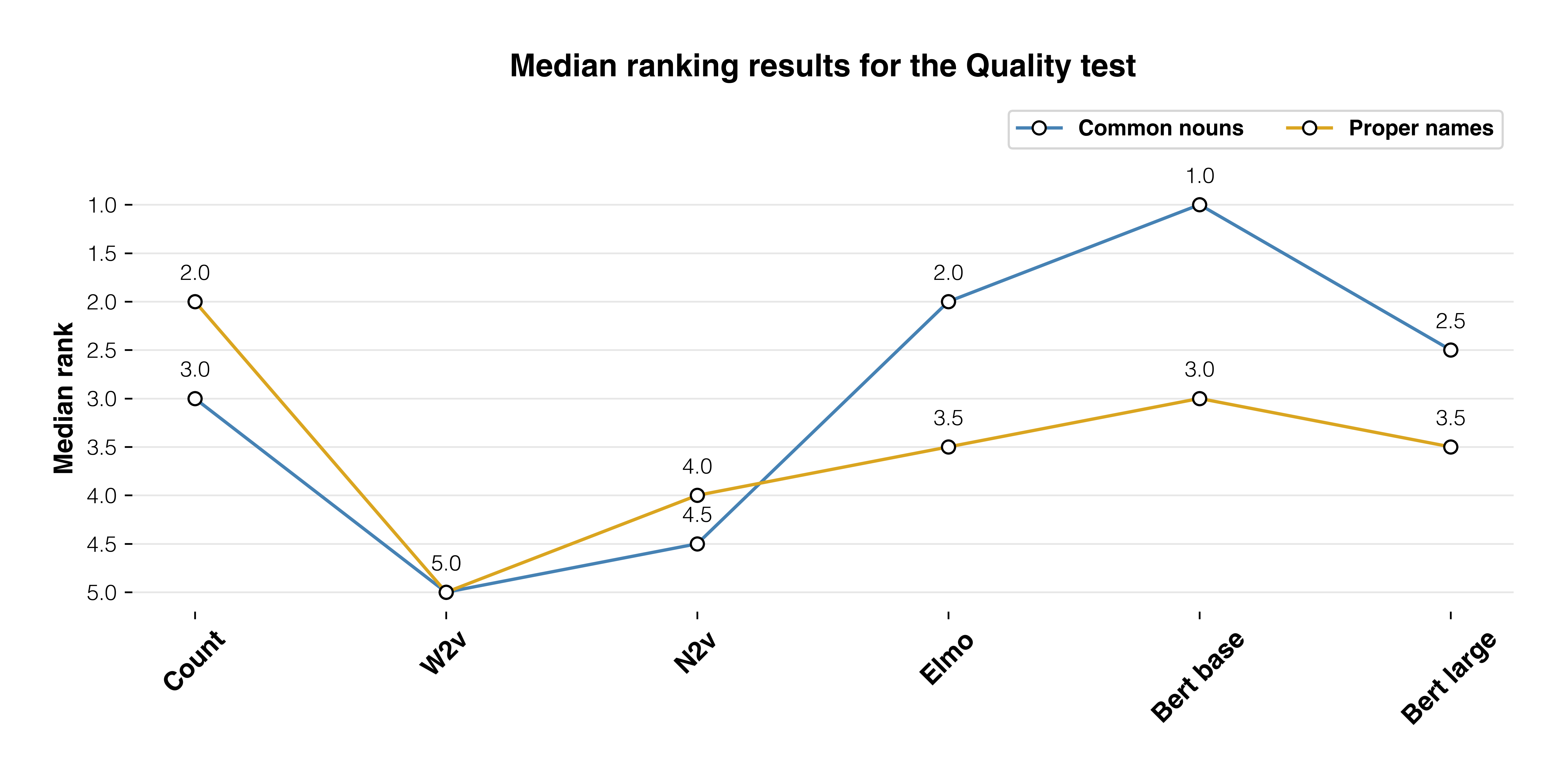}
    \caption{Results for the Quality test.}
    \label{fig:quality}
\end{figure*}

\section{Further analyses}
\input{further_analyses.tex}
\section{Conclusion}
\input{conclusion.tex}

\bibliography{bibliography}
\bibliographystyle{acl_natbib}

\appendix

\section{Distributions for the Doppelg\"{a}nger test scores} \label{appendix_A}

\begin{figure*}[ht]
    \centering
    \includegraphics[width=\textwidth]{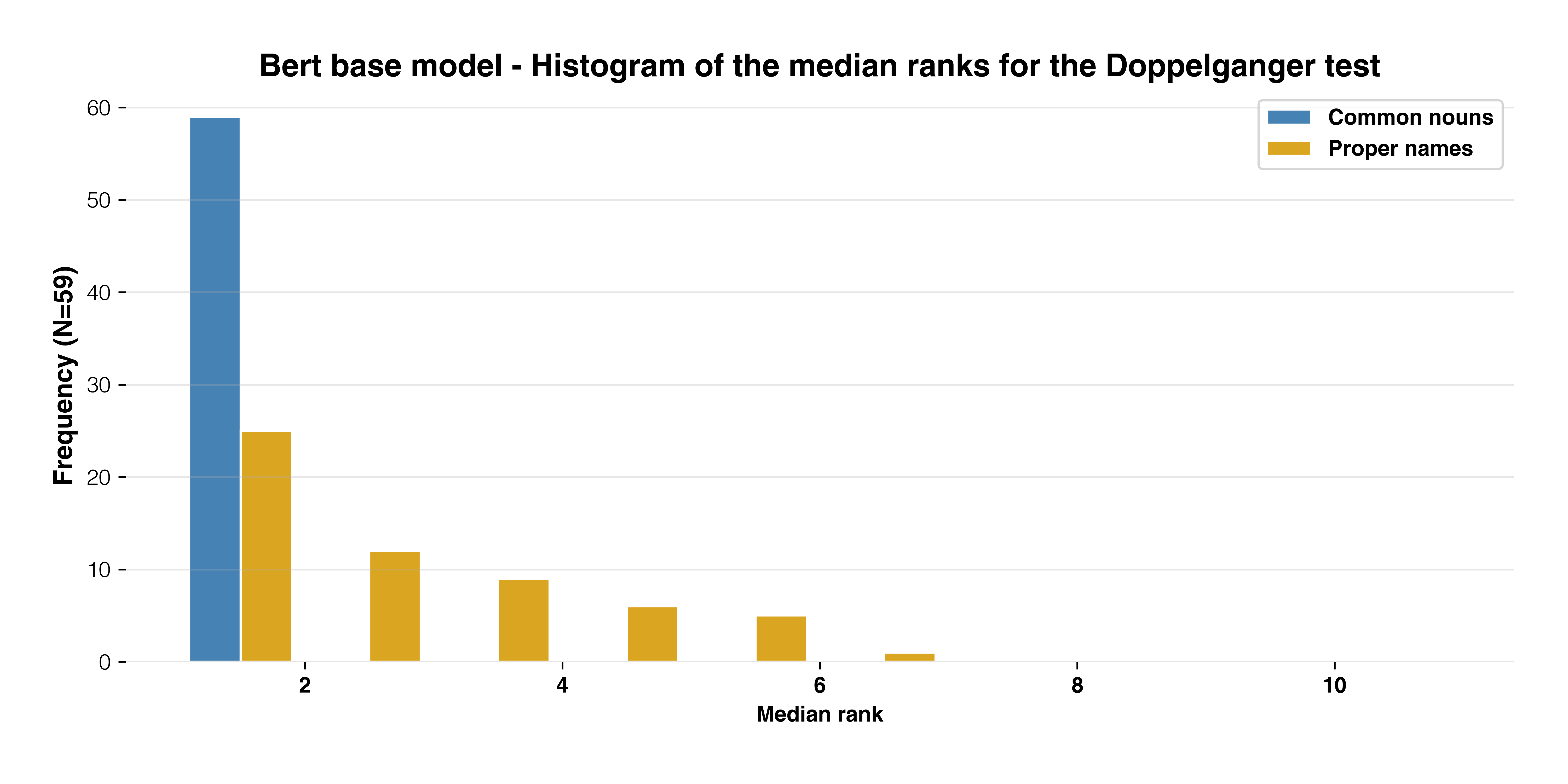}
\end{figure*}

\begin{figure*}[ht]
    \centering
    \includegraphics[width=\textwidth]{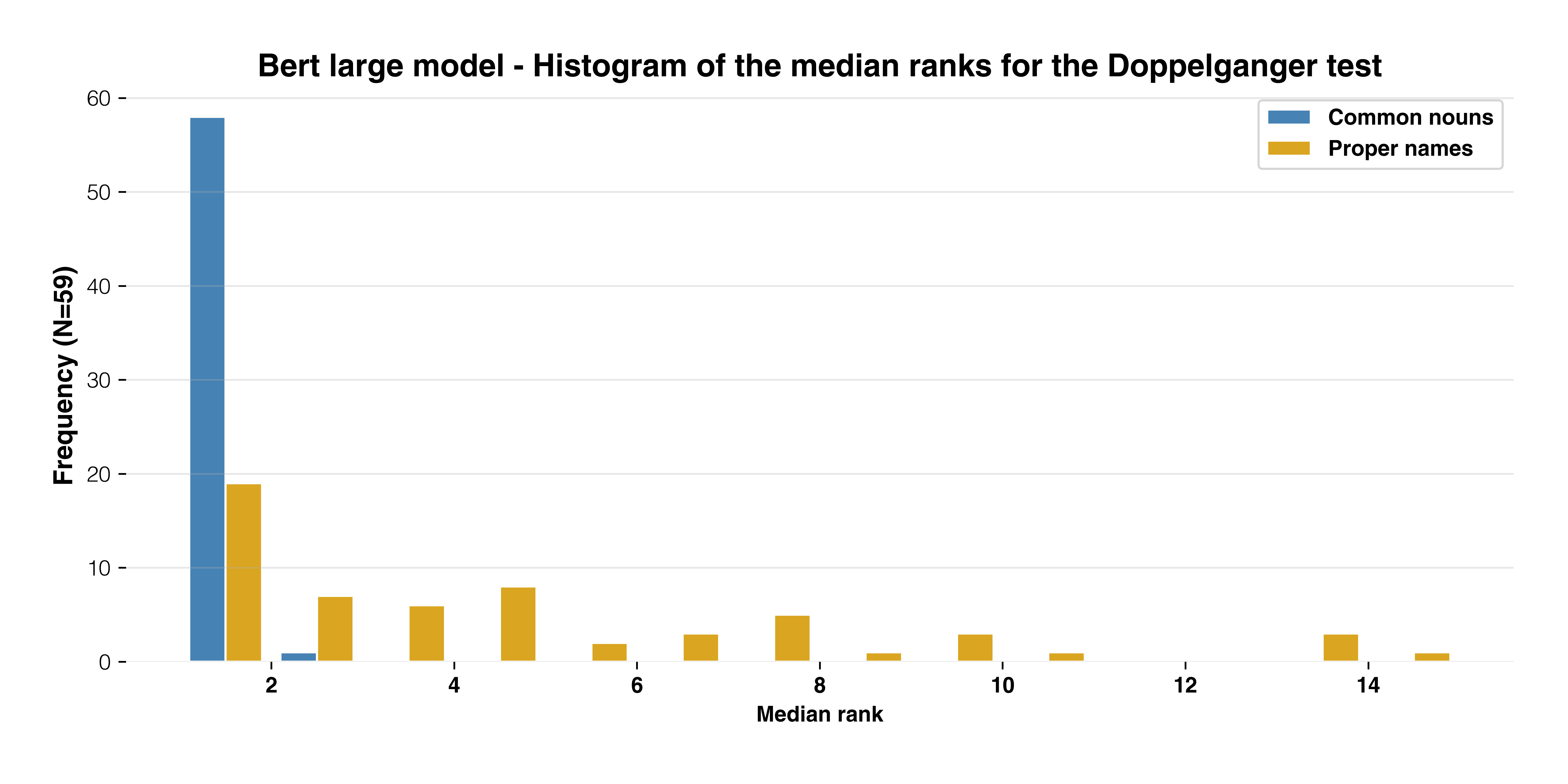}
\end{figure*}

\begin{figure*}[ht]
    \centering
    \includegraphics[width=\textwidth]{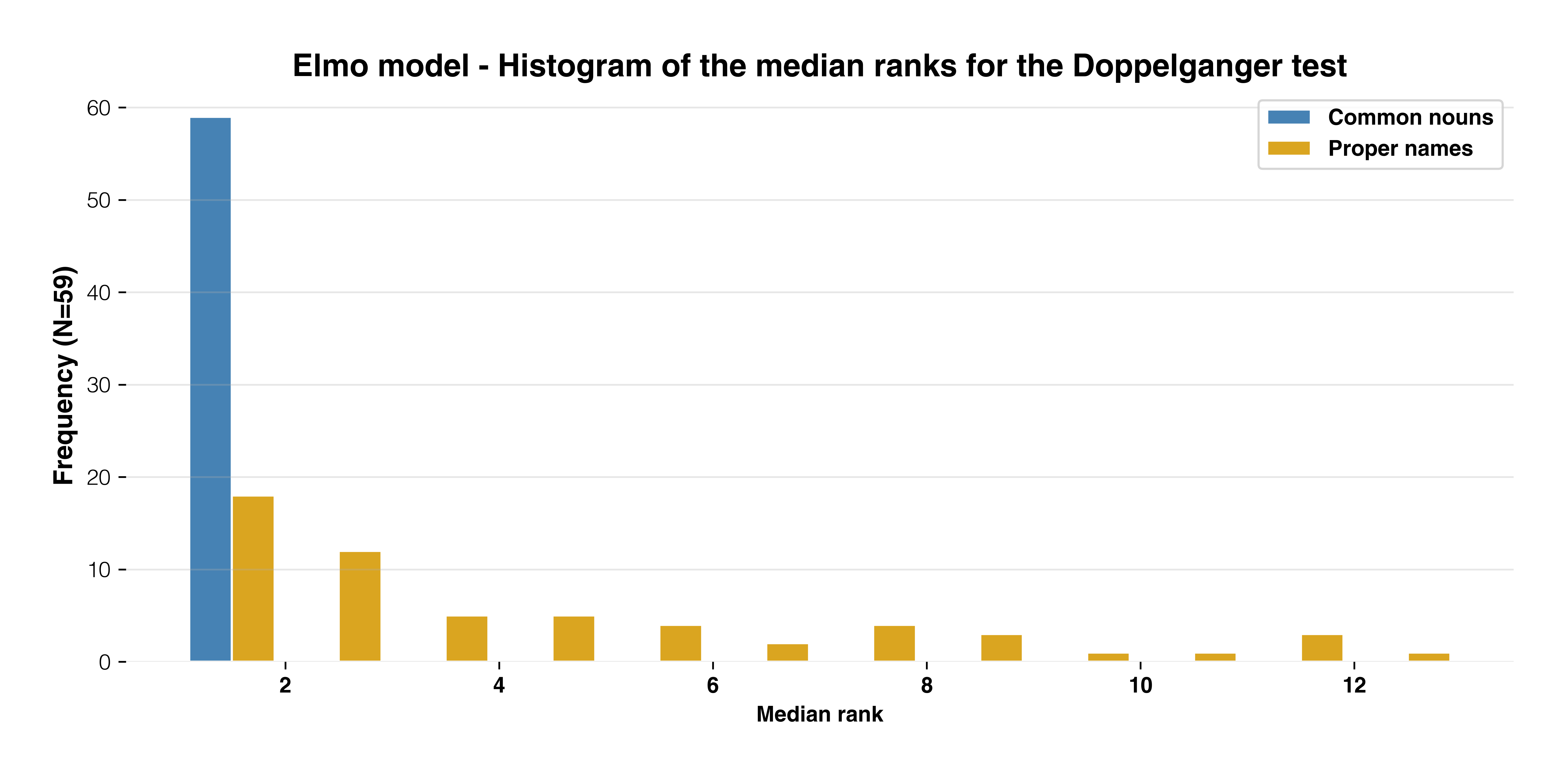}
\end{figure*}

\begin{figure*}[ht]
    \centering
    \includegraphics[width=\textwidth]{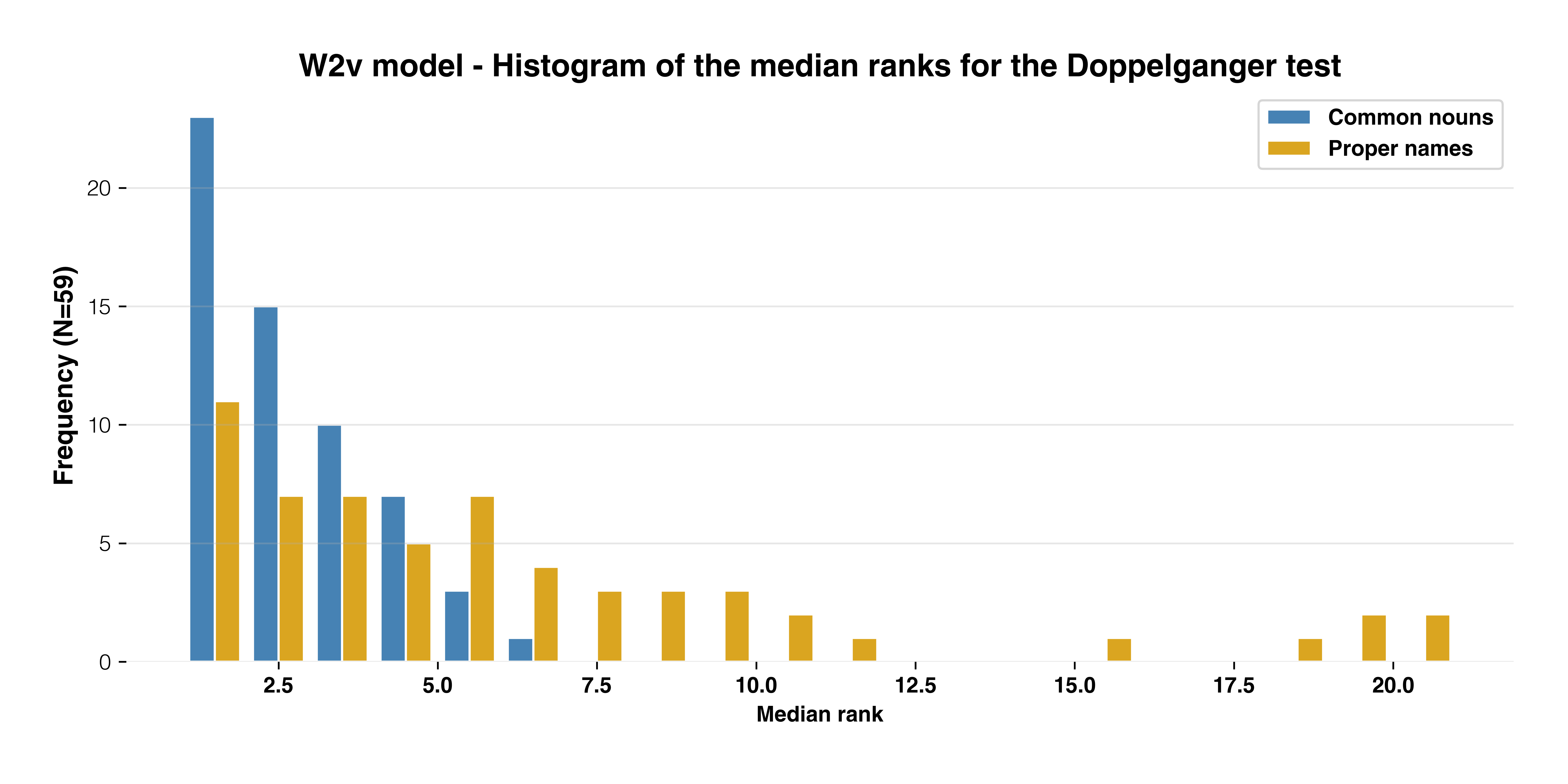}
\end{figure*}

\begin{figure*}[ht]
    \centering
    \includegraphics[width=\textwidth]{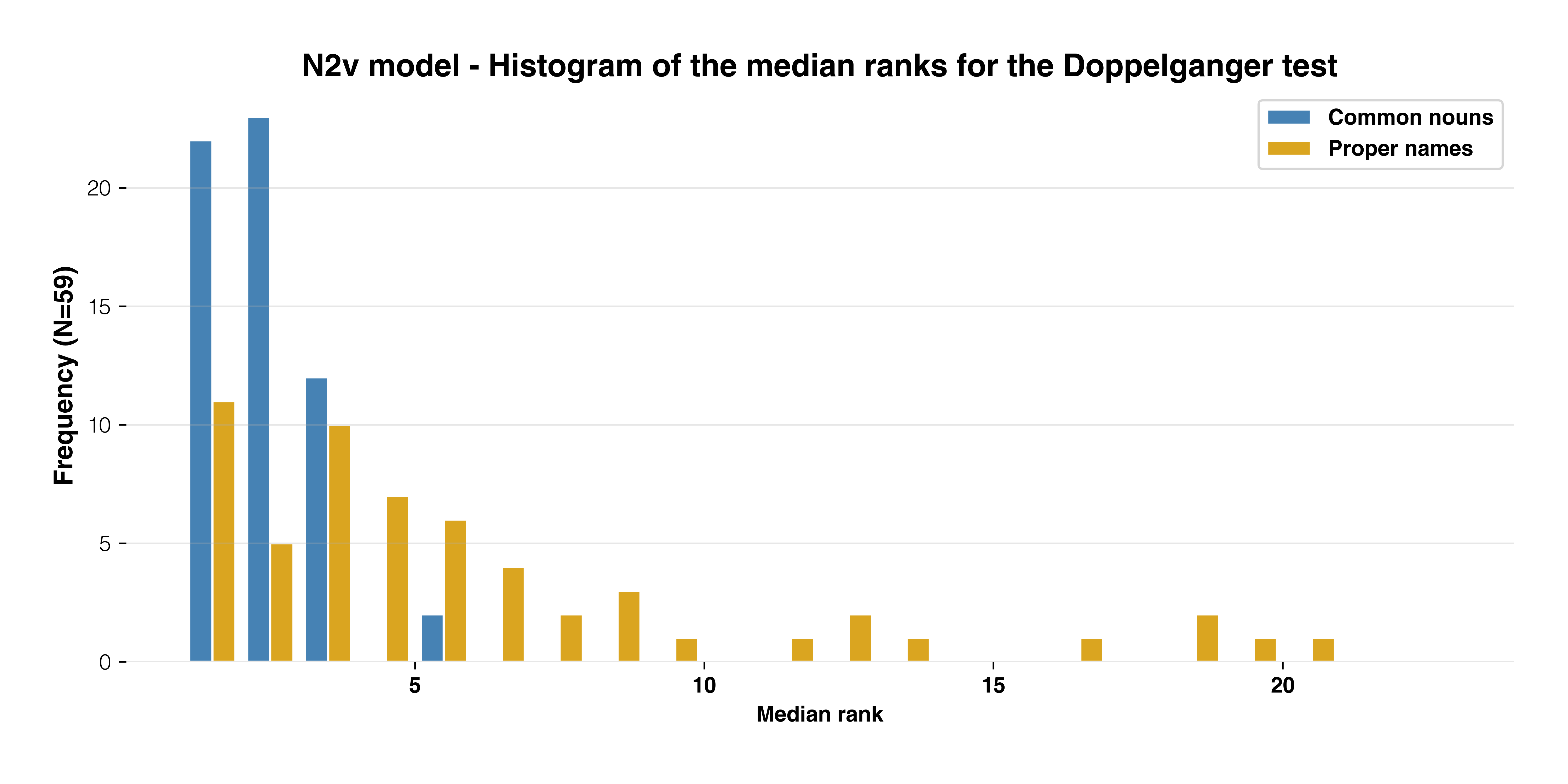}
\end{figure*}

\begin{figure*}[ht]
    \centering
    \includegraphics[width=\textwidth]{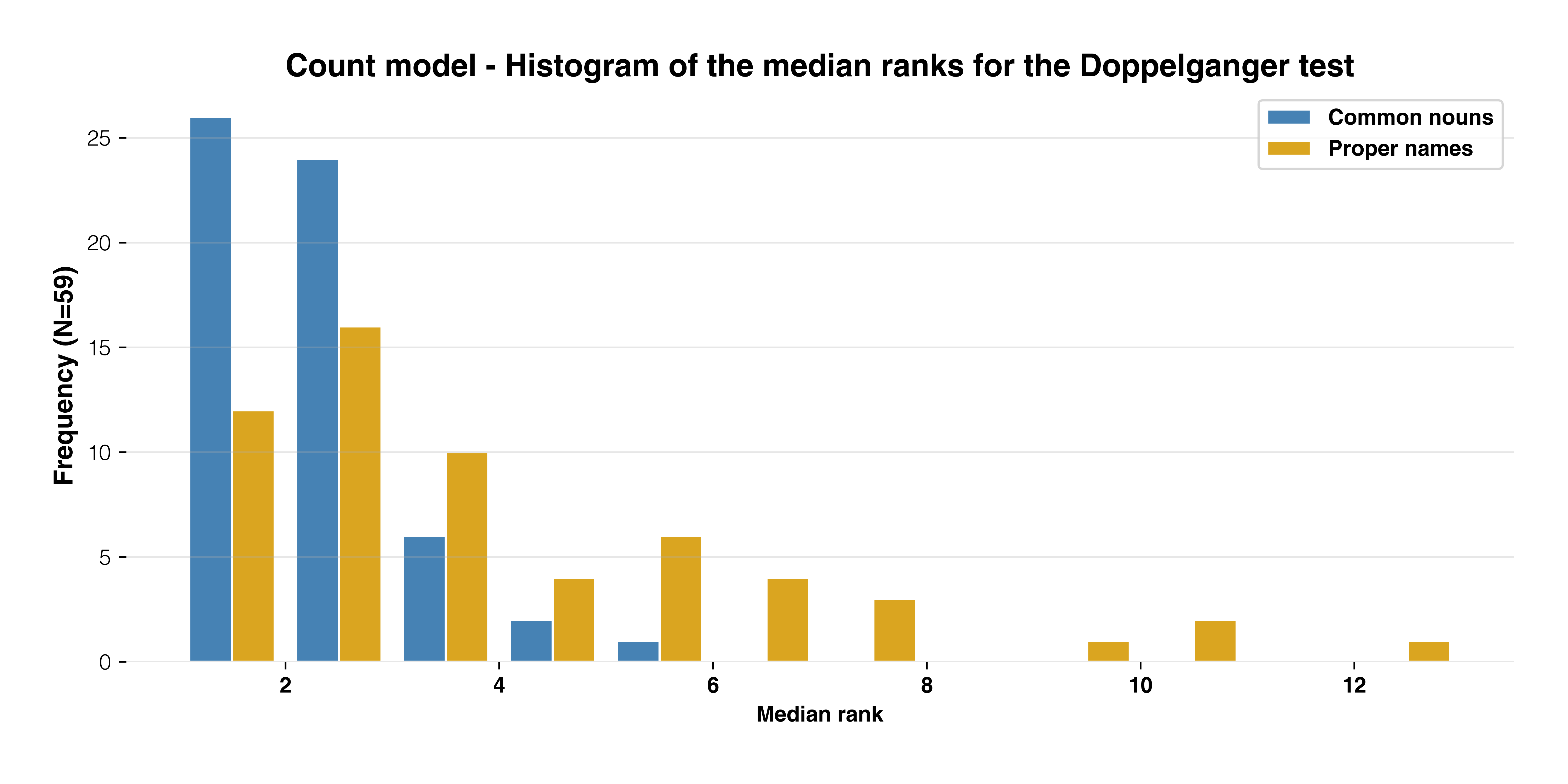}
\end{figure*}

\end{document}

%% file: abstract.tex
\begin{abstract}
In human semantic cognition, proper names (names which refer to individual entities) are harder to learn and retrieve than common nouns. This seems to be the case for machine learning algorithms too, but the linguistic and distributional reasons for this behaviour have not been investigated in depth so far. To tackle this issue, we show that the semantic distinction between proper names and common nouns is reflected in their linguistic distributions by employing an original task for distributional semantics, the Doppelgänger test, an extensive set of models, and a new dataset, the Novel Aficionados dataset. The results indicate that the distributional representations of different individual entities are less clearly distinguishable from each other than those of common nouns, an outcome which intriguingly mirrors human cognition.
\end{abstract}

%% file: introduction.tex
Learning and retrieving semantic representations for proper names is a task which, unlike other cognitive processes that are much more challenging for computers than for humans (e.g. \citealp{lake2015human}), seems difficult for both human beings (\citealp{semenza2009neuropsychology}, \citealp{bredartreview}) and machine learning algorithms (\citealp{herbelot2015}, \citealp{gupta2015}, \citealp{aina2019entity}, \citealp{almasian2019word}, \citealp{balasubramanian2020s}). Cognitive studies on the subject abound: it has been consistently found that proper names are both more difficult to acquire and retrieve from memory than common nouns and that, as a result of neurodegenerative diseases or vascular lesions, one category can be cognitively impaired independently of the other (\citealp{cohen1990}, \citealp{martins2007}). On the contrary, the linguistic properties which make proper names more difficult than common nouns for computers are a relatively unexplored field in computational linguistics and NLP. \footnote{Since names of places, objects or events have been reported in cognitive studies to dissociate from proper names of conspecifics (\citealp{lyons2002}, \citealp{crutch2004}), in order to avoid confounds, these other sorts of names won't be considered. Therefore, in the following, the expression `individual entity',  `individual' and `proper name' will be used to indicate human individuals and their names.}\newline
In contrast to common nouns, proper name representations are difficult to evaluate in computational settings \cite{chen2019enteval}. They cannot be assumed to be `known' by human annotators, so the standard evaluations (e.g. similarity, analogy) cannot be applied without extensive and costly annotation \cite{newman2018jointly}. Further, it is unclear whether such evaluations are appropriate: the meaning of a proper name is exclusively the unique individual entity it refers to, whereas common nouns refer to classes of individuals \cite{kripke1972}. So proper names are by nature extensional and should perhaps receive extensional treatment in the course of their evaluation. The main hypothesis of our work is that this difference in semantic properties between proper names and common nouns, found in human cognition, can be retrieved by distributional representations of meaning, when tested over an appropriate referential task. \newline
To show that this is the case, we propose an original referential task, the \textit{Doppelg\"{a}nger test}, associated with a new dataset, the \textit{Novel Aficionados} dataset, made of 59 novels. The Doppelg\"{a}nger test evaluates whether each entity representation learned in one subcorpus (one half of a novel) can be correctly matched to its co-referring entity representation from another subcorpus (the second half of the same novel), choosing among all the other entity representations (see figure \ref{fig:dop}). The task is challenging in that the model must distinguish between very similar entities (people and entities engaged in shared activities in a common universe) using scarce data.\newline
Using the Doppelg\"{a}nger test, we compare the distributional representations of the proper names referring to the novels' characters and those of similarly frequent common nouns mentioned in the novel. For robustness, we use several models (ELMO \citealp{peters2018deep}, BERT: \citealp{devlin2018}, Word2Vec \citealp{mikolov2013}, and Nonce2Vec \citealp{herbelotbaroni2017}). Our approach to the task is unsupervised, and in this respect it can be considered a special case of a language model probing task, focused on referential semantic information (\citealp{rogers2020primer}, \citealp{sorodoc2020probing}). By employing the same, controlled semantic representation learning procedure for proper names and common nouns within a novel, we show that distinct patterns of results for the two linguistic categories emerge.\newline
As further analyses, we look at three levels of possible discrepancies between the two categories in our setup. First, we look at low-level, distributional differences in part-of-speech neighbourhood, which confirm that they have different distributional signatures. Then, we turn to mid-level differences in terms of narrative features of the novels, with a correlational analysis. This highlights the disruptive effect of competing semantic representations, which disproportionately affect reference resolution for proper names, drawing a parallel with effects found in human semantic cognition \cite{abrams2017}. Finally, we analyze the higher-level structural differences between the obtained vector spaces, by way of a Representational Similarity Analysis study \cite{kriegeskorte2008representational} which indicates that common nouns give rise to structurally more coherent spaces than proper names. \newline
Finally, in order to show how the Doppelg\"{a}nger test can be adapted to texts from a different domains, we present the \textit{Quality test}, a challenging variation on the Doppelg\"{a}nger test which requires linking entities across different corpora (the original novels and Wikipedia).\newline 
Overall, these results suggest that proper names, when modelled by way of distributional semantics algorithms such as language models and word embeddings, require specific computational strategies in order to capture their referential properties.

%% file: related_work.tex
\subsection{Reference in NLP} The topic of proper names and that of reference have been going hand in hand in language studies for a long time, at least ever since \citet{mill1884}, \citet{frege1892} and later \citet{strawson1950} and \citet{kripke1972}. With respect to our approach, the most closely related tasks in computational linguistics are entity linking (EL) and anaphora (or co-reference) resolution. \newline
Entity Linking, also called Named Entity Disambiguation, is a NLP task where the correct reference of mentions of proper names in a text has to be found in a knowledge database (\citealp{NED}, \citealp{onoe2020fine}). Entity Linking models have no interest in modelling in any way cognitive processes (indicated for instance by the fact that they often use strong supervision), whereas our model is kept unsupervised, in order to obtain evidence which can be theoretically interpretable. \newline
Anaphora resolution is the name of the process by which a competent speaker naturally gets to understand that in the sentence ``Saul and Tina went to the market: he bought a pin and she bought a fake gun'' the word `he' refers to the same individual `Saul' refers to, and `she' refers to the referent of `Tina'. Various algorithms and tasks have been proposed in order to model this linguistic phenomenon in computational linguistics \cite{poesiobook}. However, as anaphora resolution can be modelled by employing the same strategies for both common nouns and proper names, the two linguistic categories have, to our knowledge, not been investigated separately \cite{clark2016improving}. \newline
A perspective more akin to the present one is that of \citet{herbelot2015} where, given the poor quality of distributional semantics representations of characters as extracted from two novels, the author presented ad-hoc techniques in order to improve those semantic representations. It is important to underline, however, that the focus of the present work is different from the one of \citet{herbelot2015}: here the goal is not at all that of finding ways to extract better representations for proper names in distributional semantics. Rather, the aim is that of studying, from a distributional and cognitively-oriented perspective, why proper names are more difficult than common nouns for computational semantic processing in the first place.\newline
In this sense, as it focuses on theoretical investigation, our work is more similar to \cite{gupta2015distributional} and \cite{gupta2018instantiation} which respectively try to extract attributes and categories for proper names from distributional models.\newline
\subsection{Characters in novels}
Work on individual entities - the kinds of entities proper names refer to - in computational linguistics and NLP has often made use of novels. However, such approaches have concentrated mainly on conceptually different tasks: learning character types (\citealp{bamman2013learning}, \citealp{flekova2015personality}), inferring characters' features \cite{louis2018}, relations (\citealp{iyyer2016}, \citealp{elson2010}), networks \cite{labatut2019extraction} or on broader natural language understanding tasks \cite{frermann2018}, such as inferring plot structure \cite{elsner2012character}. Here, instead, the focus is on the investigation of the semantic and referential distinction between proper names and common nouns, whose distinct categorical status is a solid cross-linguistic phenomenon \cite{vannames}. 

%% file: data.tex
In order to carry out our experiments, we collected a new dataset, the \textit{Novel Aficionados} dataset.\footnote{The dataset is available at \url{https://github.com/andreabruera/novel\_aficionados}} The core material of the dataset is made of 59 novels, collected from the Project Gutenberg website\footnote{\url{http://www.gutenberg.org/}}, an online repository of free ebooks. They were selected from the list of the 100 most downloaded ebooks of the month at the time of data collection, by excluding non-fiction ebooks. All novels are not protected by copyright anymore.\newline
Narrative literature is particularly suited to our approach, because of the importance of characters in narration. Fiction plots are built around characters, which are (often non-existing) human individuals, and around their thoughts and actions. In this sense, novels are written precisely in order to allow the creation of semantic representations of the individual entities by way of text only \cite{bamman2019annotated}.\newline
The dataset consists of an augmented and annotated version of the novels. First, all character mentions, which often take various forms despite referring to the same entity (e.g. `Mr. Darcy', `Darcy Fitzwilliam' and `Darcy'), are substituted by a unique label (in our example, `mr\_darcy') and marked by two `\$' symbols, before and after the mention (`\$mr\_darcy\$'). For the analyses, only characters occurring more than 10 times are retained. Secondly, the most frequent common nouns (considering their lemmas) for each novel are selected in order to be used for the Doppelg\"{a}nger test. Their number was matched to the amount of characters previously annotated. The rationale for choosing the most frequent common nouns as the counterpart to the characters in the dataset is that they arguably capture the novel's main themes and topic. To distinguish them from the characters' names, the selected common nouns are surrounded by two `\#' symbols (e.g. `\#hound\#').\newline
For this process of data augmentation and annotation, we used BookNLP \citep{booknlp}, a full NLP pipeline optimized for novels which importantly includes both Named Entity Recognition and co-reference resolution modules.\newline
Finally, the dataset is enriched with the matched Wikipedia pages for each one of the 59 novels, processed using the same annotation style. This data is included as it constitutes a non-narrative, encyclopedic source of information about the characters, themes and topics present in the novels. An example of use of this portion of the dataset is presented in section \ref{quality}.\newline
Each file in the dataset was split into sentences by using Spacy\footnote{\url{https://spacy.io/}}, so that each line of the resulting files contains a single sentence. Also, all punctuation was removed and the letters were turned into lower case.\newline

%% file: task.tex
The Doppelg\"{a}nger test aims at probing referential information contained in distributional semantic representations. It starts from the intuition that we should be able to match two different representations of the same referent, even if they are obtained from distinct data. In order to reduce confounds, in the Doppelg\"{a}nger test we take a single document where multiple entities appear - in our case, a novel, where entities are referred to by either proper names and common nouns. Then, the document is first split into two sub-corpora (Part A and Part B), both containing mentions of all the entities (see figure \ref{fig:dop}). Subsequently, for each part, a semantic representation for each entity is obtained by way of a distributional semantics model. Finally, by taking the two separate sets of representations, containing the same entities but coming from different parts of the document, the Doppelg\"{a}nger test probes to what extent it is possible to match the co-referring vectors with one another.\newline
It is a purely referential, extensional task, as it evaluates word vectors on the basis of their ability to model the extension (the reference) of a word. Because of this, it is naturally suited to comparing the capabilities of distributional semantic models with respect to two categories,  proper names and common nouns, whose referential properties are different: proper names refer to unique entities, whereas common nouns refer to classes of individuals. This is the question that we focus on in this work - however, the Doppelg\"{a}nger test can be used as a generic probing task for distributional semantic models, and computational models of semantics at large.\newline
In this work we decided to keep a strictly unsupervised approach to the Doppelg\"{a}nger test. This is in line with recent work in language models probing (\citealp{broscheit2019investigating}, \citealp{petroni2019language}, \citealp{talmor2020olmpics}), whose goal is to investigate as directly as possible the behaviour of the models' representations on the task at hand.\newline
More precisely, in the current setup, given a novel $N$, we split it in two halves, $N_A$ and $N_B$. We experiment with both splitting a novel in two parts at the original midpoint, and with first randomizing the list of sentences, then splitting the randomized sentences in two halves, averaging the results for 100 iterations. No difference in results emerged, so we chose the former, simpler approach. Entities present in only one of the two parts were not retained for further analyses. We found this had very little impact on the final amount of entities used. \newline
The analyses for proper names and common nouns are carried out separately, making sure to employ the same number of entities for each category. The two categories are compared only at the end. From each part, we obtain a matched set of word vectors $E_{part} = \{ \vec{e_{1}} ... \vec{e_{n}} \}$, either referring to characters or to common nouns' referents. In order to probe the performance at the Doppelg\"{a}nger test, we use a simple unsupervised ranking approach. For each vector in $E_A$ (and then conversely in $E_B$), the query $\vec{e_{A_i}}$, we compute the pairwise cosine similarities with all vectors in $E_B$, then we rank the vectors in $E_B$ according to their similarity to the query $\vec{e_{A_i}}$. The position in the ranking of the co-referring vector $\vec{e_{B_i}}$ constitutes the model's performance with respect to the current entity. The median of the per-entity scores is the per-novel score, and the median of the 59 per-novel scores constitutes the final score for the model at hand.\newline
Scores for all the models and semantic categories (common nouns or proper names) are compared in figure \ref{fig:dop_results}.

%% file: models.tex
We employed a broad range of distributional semantic models, so as to avoid biases inherent in specific implementations: they all rely on the Distributional Hypothesis \cite{firth1957synopsis}, which states that words found in similar contexts have similar meanings, but they all differ in their realization \cite{pilehvar2020embeddings}. We used three kinds of models: count-based, prediction-based (following the terminology of \citet{baroni2014}), and contextualized language models. In all models, and for each novel, both sets of vectors $E_{A}$ and $E_{B}$ are initialized as two sets of vectors filled with zeroes, and they are then updated by using the novel's data.\newline
The \textbf{count} model is based on simple word co-occurrence counts, transformed to PPMI measures, a correction which has been shown to drastically improve performances \cite{goldberg2014word2vec}. Co-occurrences were counted by considering a sliding window of 5 words to the right and to the left of each target word.\newline
The prediction-based models are \textbf{Word2Vec} (W2V), a very successful language model consisting of a feed-forward neural network \cite{mikolov2013}, and \textbf{Nonce2Vec} (N2V), a modified version of Word2Vec, specialized for small datasets such as novels \cite{herbelotbaroni2017}. First, we pre-trained a Word2Vec model on the English version of Wikipedia in its Python Gensim implementation \cite{rehurek2010software}, using the skip-gram training method and default parameters.\newline
For \textbf{Word2Vec}, each entity mention was modeled as the average of the pre-trained model's vectors for the words surrounding it, again within a window of 5 words on each side. The final set of vectors $E_{part}$ was obtained by representing each entity by the average of its mentions' vectors.\newline 
In the case of \textbf{Nonce2Vec}, the same pre-trained Word2Vec model was used. However, \textbf{Nonce2Vec} allows to adapt the skip-gram training regime to the reduced amount of data offered by a novel, thus creating new entity representations by exploiting the pre-trained weights.\newline
As contextualized models we used BERT (both BERT-BASE and BERT-LARGE \cite{devlin2018}, in their Python huggingface implementation \cite{wolf2019huggingface}) and ELMO \cite{peters2018deep}. They have been shown to have comparable performances, but they differ in several respects. In order to predict a target word, ELMO first models separately left and right context by two separate LSTMs, eventually merging their representations. BERT, instead, employs the Transformer architecture \cite{vaswani2017}, considering at the same time all the words around the target one. \newline 
In our setup, for both models, given a sentence containing an entity mention, we first mask the mention, thus making it the unknown target word to be predicted. Then we provide the full sentence, with the masked entity mention, to the model for a forward pass. Finally the vector corresponding to the masked entity mention is extracted from the last hidden layer of the contextualized language model. As in \textbf{Word2Vec}, the final representation for an entity is obtained by averaging all the vectors for its mentions.

%% file: results.tex
Results are shown in figure \ref{fig:dop_results}. All models perform better when matching representations for common nouns, than for proper names. An inspection of the distributions of the scores for each model (see appendix \ref{appendix_A}) confirms that whereas common nouns most often obtain a score of 1, indicating that the referential task was carried successfully, the distribution for proper names is much less skewed towards 1 and has a much longer tail.\newline 
This pattern of results is strikingly consistent across models, indicating that the semantic, referential distinction between proper names and common nouns emerges in the acquisition of semantic representations even when using exclusively textual, distributional linguistic information.

%% file: further_analyses.tex
In order to understand what drives such a consistent pattern of results, we carried out three separate investigations, focusing on three levels: low-level distributional features, by way of a part-of-speech neighbourhood analysis; novel-level variables such as length in words, number of characters involved and differences in characters' mentions; vector space-level analyses, by way of Representational Similarity Analysis.

\subsection{Part-of-speech neighbourhood}

To quantify the differences in the distributional properties of proper names and common nouns, we looked at the part-of-speech occurrences around the characters' names and the chosen common nouns in the Novel Aficionados dataset. We used a sliding window of 2 words on both sides of each mention, and we kept track of the co-occurrences in a matrix. The matrix had two rows, one for proper names and one for common nouns, and six columns corresponding to six parts-of-speech categories: adjectives (ADJ), adverbs (ADV), determiners (DET), nouns (NOUN), pronouns (PRON) and verb (VERB). The part-of-speech tagging was carried out with the Spacy toolkit. Aside from the obvious difference in the frequency of determiners (in English proper names can't have a determiner before them), this analysis shows that proper names are more frequently found in the vicinity of nouns and verbs than common nouns, confirming that there are low-level differences in surrounding word distributions amongst the two categories. 

\subsection{Correlational analysis}

Novels may be characterized by structural features which make it more difficult to match co-referring word vectors for characters: some novels may be very short, thus not providing enough data; some may have a larger amount of characters, making it more difficult to correctly discriminate among different characters' representations; finally, in some novels some characters may receive much more attention than others, a case of uneven data split which may affect results. In order to understand the importance of these variables in the Doppelg\"{a}nger results, we looked at the correlation between the models' scores and the three variables (novel length, number of characters, standard deviation of mentions across characters). Results are shown in figure \ref{fig:correlations}. Both novel length and number of characters correlate strongly with results, with the latter dominating in all models. This entails that, as the number of characters increases, the representations for the characters get progressively confused with one another, and that distributional models have a hard time with correctly establishing reference for numerous entities. This result dovetails with both cognitive \cite{abrams2017} and computational findings \cite{ilievski2018systematic}.

\subsection{Representational similarity analysis}

Finally, for each novel, we compared the properties of the vector spaces corresponding to the two portions of the novels. We wanted to find out whether the resulting vector spaces across the two parts of the novels, $E_A$ and $E_B$, were significantly different in their structural properties between proper names and common nouns. An ideal framework to carry out such analyses is that of Representational Similarity Analysis \cite{kriegeskorte2008representational}, originally proposed in cognitive neuroscience. \newline 
In this approach two different vector spaces (having the same, matched amount of vectors) are not compared directly, but rather by way of the vectors of their within-space pairwise similarities. These two pairwise similarity vectors encode the representational structure of each space; and if two such vectors correlate, then they are taken to be similar in their representational structure. For each novel, and separately for each word category, we look at the correlation of the matched pairwise similarities for the two vector spaces $E_A$ and $E_B$. Results are shown in figure \ref{fig:rsa}.\newline
Proper names exhibit lower representational similarities across vector spaces in all models. It seems reasonable to speculate that this structural difference must play an important role in the Doppelg\"{a}nger scores, where structurally less similar pairs of vector spaces (those for proper names) perform worse.

\subsection{Going beyond the novels: the Quality test} \label{quality}

It is important to understand whether our results are specific to novels, or can generalize to other domains and kinds of text. As a first step, we include a different implementation of the Doppelg\"{a}nger test, that we call the \textit{Quality test}. In this test, for each set of entities, instead of using two sub-documents, we use two different kinds of documents: a novel and a Wikipedia page on the novel, which is included in the Novel Aficionados dataset for each novel.\newline
The Wikipedia description of a novel includes information about the same characters and entities as the novel itself, but it presents it with both a different purpose (short presentation of fundamental features) and a different style (non-narrative). Therefore, the task should be more difficult than the original Doppelg\"{a}nger test, because of the added difficulty due to the difference between the two documents used for creating the sets of entity representations $E_{A}$ and $E_{B}$.\newline
As it can be seen from figure \ref{fig:quality}, results have a partially different pattern with respect to the Doppelg\"{a}nger test. Contextualized models perform similarly to the original test, confirming their ability to encode semantic information solidly, even in challenging conditions. In this case too, contextualized models show worse performance for proper names. This confirms that this semantic category poses peculiar challenges to distributional semantic models. The models based on Word2Vec models perform very poorly, and on a par for proper names and common nouns, indicating that they are not very robust to this experimental manipulation. Finally, the performance for the count-based model show the reversed pattern: a puzzling result which calls for an application of the Doppelg\"{a}nger test to different types of texts.

%% file: conclusion.tex
Using as a starting point the distinction between proper names and common nouns, fairly well studied in the neuro-cognitive and formal semantics literature, but almost ignored in computational linguistics and NLP, we proposed a new evaluation for computational representations of entities, the Doppelg\"{a}nger test. This task probes in particular for referential, extensional semantic information encoded in those representations, which is of paramount importance specifically for proper names. It does so by first splitting a document into two sub-documents, then obtaining two matched sets of semantic representations for the entities contained in the document, and finally evaluating to what extent it is possible to match the pairs of co-referring vectors. \newline 
We compared the performances of an extensive set of distributional semantic models by using an original dataset, the \textit{Novel Aficionados} dataset, tailored to comparing the models' performances on proper names and common nouns. By means of the Doppelg\"{a}nger test the semantic distinction between the two categories emerged in strikingly different patterns of results. What's more, the models' performances mirrored human cognition, with common nouns being consistently easier to match according to their reference than proper names.\newline 
By way of further analyses, we showed that the distinction between the two categories is present both at the level of textual distributional properties, in the form of part-of-speech co-occurrence differences, and at the level of vector space structure, which is more similar across matched sets of vectors for common nouns than proper names. Also, models were shown to be gradually degrading their performance as more individual entities were considered. \newline
Finally, by using the Doppelg\"{a}nger test on different data, we demonstrated how it can become, beyond the current setup, a valuable evaluation framework for probing for referential information in semantic representations of individual entities.

%% file: main.bbl
\begin{thebibliography}{54}
\expandafter\ifx\csname natexlab\endcsname\relax\def\natexlab#1{#1}\fi

\bibitem[{Abrams and Davis(2017)}]{abrams2017}
Lise Abrams and Danielle~K Davis. 2017.
\newblock Competitors or teammates: how proper names influence each other.
\newblock \emph{Current Directions in Psychological Science}, 26(1):87--93.

\bibitem[{Aina et~al.(2019)Aina, Silberer, Sorodoc, Westera, and
  Boleda}]{aina2019entity}
Laura Aina, Carina Silberer, Ionut Sorodoc, Matthijs Westera, and Gemma Boleda.
  2019.
\newblock What do entity-centric models learn? insights from {Entity} {Linking}
  in multi-party dialogue.
\newblock In \emph{Proceedings of the 2019 Conference of the North American
  Chapter of the Association for Computational Linguistics: Human Language
  Technologies, Volume 1 (Long and Short Papers)}, pages 3772--3783.

\bibitem[{Almasian et~al.(2019)Almasian, Spitz, and Gertz}]{almasian2019word}
Satya Almasian, Andreas Spitz, and Michael Gertz. 2019.
\newblock Word embeddings for entity-annotated texts.
\newblock In \emph{European Conference on Information Retrieval}, pages
  307--322. Springer.

\bibitem[{Balasubramanian et~al.(2020)Balasubramanian, Jain, Jindal, Awasthi,
  and Sarawagi}]{balasubramanian2020s}
Sriram Balasubramanian, Naman Jain, Gaurav Jindal, Abhijeet Awasthi, and Sunita
  Sarawagi. 2020.
\newblock What’s in a name? are bert named entity representations just as
  good for any other name?
\newblock In \emph{Proceedings of the 5th Workshop on Representation Learning
  for NLP}, pages 205--214.

\bibitem[{Balog(2018)}]{NED}
Krisztian Balog. 2018.
\newblock \emph{Entity {Linking}}, pages 147--188. Springer International
  Publishing, Cham.

\bibitem[{Bamman et~al.(2013)Bamman, O’Connor, and
  Smith}]{bamman2013learning}
David Bamman, Brendan O’Connor, and Noah~A Smith. 2013.
\newblock Learning latent personas of film characters.
\newblock In \emph{Proceedings of the 51st Annual Meeting of the Association
  for Computational Linguistics (Volume 1: Long Papers)}, pages 352--361.

\bibitem[{Bamman et~al.(2019)Bamman, Popat, and Shen}]{bamman2019annotated}
David Bamman, Sejal Popat, and Sheng Shen. 2019.
\newblock An annotated dataset of literary entities.
\newblock In \emph{Proceedings of the 2019 Conference of the North American
  Chapter of the Association for Computational Linguistics: Human Language
  Technologies, Volume 1 (Long and Short Papers)}, pages 2138--2144.

\bibitem[{Bamman et~al.(2014)Bamman, Underwood, and Smith}]{booknlp}
David Bamman, Ted Underwood, and Noah~A Smith. 2014.
\newblock A bayesian mixed effects model of literary character.
\newblock In \emph{Proceedings of the 52nd Annual Meeting of the Association
  for Computational Linguistics (Volume 1: Long Papers)}, volume~1, pages
  370--379.

\bibitem[{Baroni et~al.(2014)Baroni, Dinu, and Kruszewski}]{baroni2014}
Marco Baroni, Georgiana Dinu, and Germ{\'a}n Kruszewski. 2014.
\newblock Don't count, predict! {A} systematic comparison of context-counting
  vs. context-predicting semantic vectors.
\newblock In \emph{Proceedings of the 52nd Annual Meeting of the Association
  for Computational Linguistics (Volume 1: Long Papers)}, volume~1, pages
  238--247.

\bibitem[{Br{\'e}dart(2017)}]{bredartreview}
Serge Br{\'e}dart. 2017.
\newblock The cognitive psychology and neuroscience of naming people.
\newblock \emph{Neuroscience \& Biobehavioral Reviews}, 83:145--154.

\bibitem[{Broscheit(2019)}]{broscheit2019investigating}
Samuel Broscheit. 2019.
\newblock Investigating entity knowledge in bert with simple neural end-to-end
  entity linking.
\newblock In \emph{Proceedings of the 23rd Conference on Computational Natural
  Language Learning (CoNLL)}, pages 677--685.

\bibitem[{Chen et~al.(2019)Chen, Chu, Chen, Stratos, and
  Gimpel}]{chen2019enteval}
Mingda Chen, Zewei Chu, Yang Chen, Karl Stratos, and Kevin Gimpel. 2019.
\newblock Enteval: A holistic evaluation benchmark for entity representations.
\newblock In \emph{Proceedings of the 2019 Conference on Empirical Methods in
  Natural Language Processing and the 9th International Joint Conference on
  Natural Language Processing (EMNLP-IJCNLP)}, pages 421--433.

\bibitem[{Clark and Manning(2016)}]{clark2016improving}
Kevin Clark and Christopher~D Manning. 2016.
\newblock Improving coreference resolution by learning entity-level distributed
  representations.
\newblock In \emph{Proceedings of the 54th Annual Meeting of the Association
  for Computational Linguistics (Volume 1: Long Papers)}, pages 643--653.

\bibitem[{Cohen(1990)}]{cohen1990}
Gillian Cohen. 1990.
\newblock Why is it difficult to put names to faces?
\newblock \emph{British Journal of Psychology}, 81(3):287--297.

\bibitem[{Crutch and Warrington(2004)}]{crutch2004}
Sebastian~J Crutch and Elizabeth~K Warrington. 2004.
\newblock The semantic organisation of proper nouns: the case of people and
  brand names.
\newblock \emph{Neuropsychologia}, 42(5):584--596.

\bibitem[{Devlin et~al.(2018)Devlin, Chang, Lee, and Toutanova}]{devlin2018}
Jacob Devlin, Ming-Wei Chang, Kenton Lee, and Kristina Toutanova. 2018.
\newblock Bert: Pre-training of deep bidirectional transformers for language
  understanding.
\newblock \emph{arXiv preprint arXiv:1810.04805}.

\bibitem[{Elsner(2012)}]{elsner2012character}
Micha Elsner. 2012.
\newblock Character-based kernels for novelistic plot structure.
\newblock In \emph{Proceedings of the 13th Conference of the European Chapter
  of the Association for Computational Linguistics}, pages 634--644.
  Association for Computational Linguistics.

\bibitem[{Elson et~al.(2010)Elson, Dames, and McKeown}]{elson2010}
David~K Elson, Nicholas Dames, and Kathleen~R McKeown. 2010.
\newblock Extracting social networks from literary fiction.
\newblock In \emph{Proceedings of the 48th annual meeting of the association
  for computational linguistics}, pages 138--147. Association for Computational
  Linguistics.

\bibitem[{Firth(1957)}]{firth1957synopsis}
John~R Firth. 1957.
\newblock A synopsis of linguistic theory, 1930-1955.
\newblock \emph{Studies in linguistic analysis}.

\bibitem[{Flekova and Gurevych(2015)}]{flekova2015personality}
Lucie Flekova and Iryna Gurevych. 2015.
\newblock Personality profiling of fictional characters using sense-level links
  between lexical resources.
\newblock In \emph{Proceedings of the 2015 Conference on Empirical Methods in
  Natural Language Processing}, pages 1805--1816.

\bibitem[{Frege(1892)}]{frege1892}
Gottlob Frege. 1892.
\newblock Ueber {Sinn} und {Bedeutung}.
\newblock \emph{Zeitschrift f{\"u}r Philosophie und philosophische Kritik},
  100:25--50.

\bibitem[{Frermann et~al.(2018)Frermann, Cohen, and Lapata}]{frermann2018}
Lea Frermann, Shay~B Cohen, and Mirella Lapata. 2018.
\newblock Whodunnit? {Crime} drama as a case for natural language
  understanding.
\newblock \emph{Transactions of the Association of Computational Linguistics},
  6:1--15.

\bibitem[{Goldberg and Levy(2014)}]{goldberg2014word2vec}
Yoav Goldberg and Omer Levy. 2014.
\newblock word2vec explained: deriving {Mikolov} et al.'s negative-sampling
  word-embedding method.
\newblock \emph{arXiv preprint arXiv:1402.3722}.

\bibitem[{Gupta et~al.(2015{\natexlab{a}})Gupta, Boleda, Baroni, and
  Pad{\'o}}]{gupta2015}
Abhijeet Gupta, Gemma Boleda, Marco Baroni, and Sebastian Pad{\'o}.
  2015{\natexlab{a}}.
\newblock Distributional vectors encode referential attributes.
\newblock In \emph{Proceedings of the 2015 Conference on Empirical Methods in
  Natural Language Processing}, pages 12--21.

\bibitem[{Gupta et~al.(2015{\natexlab{b}})Gupta, Boleda, Baroni, and
  Pad{\'o}}]{gupta2015distributional}
Abhijeet Gupta, Gemma Boleda, Marco Baroni, and Sebastian Pad{\'o}.
  2015{\natexlab{b}}.
\newblock Distributional vectors encode referential attributes.
\newblock In \emph{Proceedings of the 2015 Conference on Empirical Methods in
  Natural Language Processing}, pages 12--21.

\bibitem[{Gupta et~al.(2018)Gupta, Boleda, and Pado}]{gupta2018instantiation}
Abhijeet Gupta, Gemma Boleda, and Sebastian Pado. 2018.
\newblock Instantiation.
\newblock \emph{arXiv preprint arXiv:1808.01662}.

\bibitem[{Herbelot(2015)}]{herbelot2015}
Aur{\'e}lie Herbelot. 2015.
\newblock Mr {Darcy} and {Mr} {Toad}, gentlemen: distributional names and their
  kinds.
\newblock In \emph{Proceedings of the 11th International Conference on
  Computational Semantics}, pages 151--161.

\bibitem[{Herbelot and Baroni(2017)}]{herbelotbaroni2017}
Aur{\'e}lie Herbelot and Marco Baroni. 2017.
\newblock High-risk learning: acquiring new word vectors from tiny data.
\newblock In \emph{Proceedings of the 2017 Conference on Empirical Methods in
  Natural Language Processing}, pages 304--309.

\bibitem[{Ilievski et~al.(2018)Ilievski, Vossen, and
  Schlobach}]{ilievski2018systematic}
Filip Ilievski, Piek Vossen, and Stefan Schlobach. 2018.
\newblock Systematic study of long tail phenomena in entity linking.
\newblock In \emph{Proceedings of the 27th International Conference on
  Computational Linguistics}, pages 664--674.

\bibitem[{Iyyer et~al.(2016)Iyyer, Guha, Chaturvedi, Boyd-Graber, and
  Daum{\'e}~III}]{iyyer2016}
Mohit Iyyer, Anupam Guha, Snigdha Chaturvedi, Jordan Boyd-Graber, and Hal
  Daum{\'e}~III. 2016.
\newblock Feuding families and former friends: Unsupervised learning for
  dynamic fictional relationships.
\newblock In \emph{Proceedings of the 2016 Conference of the North American
  Chapter of the Association for Computational Linguistics: Human Language
  Technologies}, pages 1534--1544.

\bibitem[{Kriegeskorte et~al.(2008)Kriegeskorte, Mur, and
  Bandettini}]{kriegeskorte2008representational}
Nikolaus Kriegeskorte, Marieke Mur, and Peter~A Bandettini. 2008.
\newblock Representational similarity analysis-connecting the branches of
  systems neuroscience.
\newblock \emph{Frontiers in systems neuroscience}, 2:4.

\bibitem[{Kripke(1972)}]{kripke1972}
Saul~A Kripke. 1972.
\newblock Naming and necessity.
\newblock In \emph{Semantics of natural language}, pages 253--355. Springer.

\bibitem[{Labatut and Bost(2019)}]{labatut2019extraction}
Vincent Labatut and Xavier Bost. 2019.
\newblock Extraction and analysis of fictional character networks: A survey.
\newblock \emph{ACM Computing Surveys (CSUR)}, 52(5):1--40.

\bibitem[{Lake et~al.(2015)Lake, Salakhutdinov, and Tenenbaum}]{lake2015human}
Brenden~M Lake, Ruslan Salakhutdinov, and Joshua~B Tenenbaum. 2015.
\newblock Human-level concept learning through probabilistic program induction.
\newblock \emph{Science}, 350(6266):1332--1338.

\bibitem[{Louis and Sutton(2018)}]{louis2018}
Annie Louis and Charles Sutton. 2018.
\newblock Deep {Dungeons} and {Dragons}: Learning character-action interactions
  from role-playing game transcripts.
\newblock In \emph{Proceedings of the 2018 Conference of the North American
  Chapter of the Association for Computational Linguistics: Human Language
  Technologies, Volume 2 (Short Papers)}, pages 708--713.

\bibitem[{Lyons et~al.(2002)Lyons, Hanley, and Kay}]{lyons2002}
Frances Lyons, J~Richard Hanley, and Janice Kay. 2002.
\newblock Anomia for common names and geographical names with preserved
  retrieval of names of people: A semantic memory disorder.
\newblock \emph{Cortex}, 38(1):23--35.

\bibitem[{Martins and Farrajota(2007)}]{martins2007}
Isabel~Pav{\~a}o Martins and Luisa Farrajota. 2007.
\newblock Proper and common names: A double dissociation.
\newblock \emph{Neuropsychologia}, 45(8):1744--1756.

\bibitem[{Mikolov et~al.(2013)Mikolov, Chen, Corrado, and Dean}]{mikolov2013}
Tomas Mikolov, Kai Chen, Greg Corrado, and Jeffrey Dean. 2013.
\newblock Efficient estimation of word representations in vector space.
\newblock \emph{arXiv preprint arXiv:1301.3781}.

\bibitem[{Mill(1884)}]{mill1884}
John~Stuart Mill. 1884.
\newblock \emph{A system of logic, ratiocinative and inductive: Being a
  connected view of the principles of evidence and the methods of scientific
  investigation}, volume~1.
\newblock Longmans, Green, and Company.

\bibitem[{Newman-Griffis et~al.(2018)Newman-Griffis, Lai, and
  Fosler-Lussier}]{newman2018jointly}
Denis Newman-Griffis, Albert~M Lai, and Eric Fosler-Lussier. 2018.
\newblock Jointly embedding entities and text with distant supervision.
\newblock In \emph{Proceedings of The Third Workshop on Representation Learning
  for NLP}, pages 195--206.

\bibitem[{Onoe and Durrett(2020)}]{onoe2020fine}
Yasumasa Onoe and Greg Durrett. 2020.
\newblock Fine-grained entity typing for domain independent entity linking.
\newblock In \emph{Proceedings of the AAAI Conference on Artificial
  Intelligence}.

\bibitem[{Peters et~al.(2018)Peters, Neumann, Iyyer, Gardner, Clark, Lee, and
  Zettlemoyer}]{peters2018deep}
Matthew Peters, Mark Neumann, Mohit Iyyer, Matt Gardner, Christopher Clark,
  Kenton Lee, and Luke Zettlemoyer. 2018.
\newblock Deep contextualized word representations.
\newblock In \emph{Proceedings of the 2018 Conference of the North American
  Chapter of the Association for Computational Linguistics: Human Language
  Technologies, Volume 1 (Long Papers)}, pages 2227--2237.

\bibitem[{Petroni et~al.(2019)Petroni, Rockt{\"a}schel, Riedel, Lewis, Bakhtin,
  Wu, and Miller}]{petroni2019language}
Fabio Petroni, Tim Rockt{\"a}schel, Sebastian Riedel, Patrick Lewis, Anton
  Bakhtin, Yuxiang Wu, and Alexander Miller. 2019.
\newblock Language models as knowledge bases?
\newblock In \emph{Proceedings of the 2019 Conference on Empirical Methods in
  Natural Language Processing and the 9th International Joint Conference on
  Natural Language Processing (EMNLP-IJCNLP)}, pages 2463--2473.

\bibitem[{Pilehvar and Camacho-Collados(2020)}]{pilehvar2020embeddings}
Mohammad~Taher Pilehvar and Jose Camacho-Collados. 2020.
\newblock Embeddings in natural language processing: Theory and advances in
  vector representations of meaning.
\newblock \emph{Synthesis Lectures on Human Language Technologies},
  13(4):1--175.

\bibitem[{Poesio et~al.(2016)Poesio, Stuckardt, and Versley}]{poesiobook}
Massimo Poesio, Roland Stuckardt, and Yannick Versley. 2016.
\newblock \emph{Anaphora Resolution}.
\newblock Springer.

\bibitem[{Rehurek and Sojka(2010)}]{rehurek2010software}
Radim Rehurek and Petr Sojka. 2010.
\newblock Software framework for topic modelling with large corpora.
\newblock In \emph{In Proceedings of the LREC 2010 Workshop on New Challenges
  for NLP Frameworks}. Citeseer.

\bibitem[{Rogers et~al.(2020)Rogers, Kovaleva, and
  Rumshisky}]{rogers2020primer}
Anna Rogers, Olga Kovaleva, and Anna Rumshisky. 2020.
\newblock A primer in bertology: What we know about how bert works.
\newblock \emph{Transactions of the Association for Computational Linguistics},
  8:842--866.

\bibitem[{Semenza(2009)}]{semenza2009neuropsychology}
Carlo Semenza. 2009.
\newblock The neuropsychology of proper names.
\newblock \emph{Mind \& Language}, 24(4):347--369.

\bibitem[{Sorodoc et~al.(2020)Sorodoc, Gulordava, and
  Boleda}]{sorodoc2020probing}
Ionut Sorodoc, Kristina Gulordava, and Gemma Boleda. 2020.
\newblock Probing for referential information in language models.
\newblock In \emph{Proceedings of the 58th Annual Meeting of the Association
  for Computational Linguistics}, pages 4177--4189.

\bibitem[{Strawson(1950)}]{strawson1950}
Peter~F Strawson. 1950.
\newblock On referring.
\newblock \emph{Mind}, 59(235):320--344.

\bibitem[{Talmor et~al.(2020)Talmor, Elazar, Goldberg, and
  Berant}]{talmor2020olmpics}
Alon Talmor, Yanai Elazar, Yoav Goldberg, and Jonathan Berant. 2020.
\newblock olmpics-on what language model pre-training captures.
\newblock \emph{Transactions of the Association for Computational Linguistics},
  8:743--758.

\bibitem[{Van~Langendonck and Van~de Velde(2016)}]{vannames}
Willy Van~Langendonck and Mark Van~de Velde. 2016.
\newblock Names and grammar.
\newblock In \emph{The Oxford Handbook of Names and Naming}. Oxford University
  Press.

\bibitem[{Vaswani et~al.(2017)Vaswani, Shazeer, Parmar, Uszkoreit, Jones,
  Gomez, Kaiser, and Polosukhin}]{vaswani2017}
Ashish Vaswani, Noam Shazeer, Niki Parmar, Jakob Uszkoreit, Llion Jones,
  Aidan~N Gomez, {\L}ukasz Kaiser, and Illia Polosukhin. 2017.
\newblock Attention is all you need.
\newblock In \emph{Advances in neural information processing systems}, pages
  5998--6008.

\bibitem[{Wolf et~al.(2019)Wolf, Debut, Sanh, Chaumond, Delangue, Moi, Cistac,
  Rault, Louf, Funtowicz et~al.}]{wolf2019huggingface}
Thomas Wolf, Lysandre Debut, Victor Sanh, Julien Chaumond, Clement Delangue,
  Anthony Moi, Pierric Cistac, Tim Rault, R{\'e}mi Louf, Morgan Funtowicz,
  et~al. 2019.
\newblock Huggingface's transformers: State-of-the-art natural language
  processing.
\newblock \emph{arXiv preprint arXiv:1910.03771}.

\end{thebibliography}
